\documentclass{article}

\usepackage{arxiv}

\usepackage{graphicx}
\usepackage[utf8]{inputenc} 
\usepackage[T1]{fontenc}    
\usepackage{hyperref}       
\usepackage{url}            
\usepackage{booktabs}       
\usepackage{amsfonts}       
\usepackage{nicefrac}       
\usepackage{microtype}      
\usepackage{lipsum}
\usepackage{amsmath}
\usepackage{cases}
\usepackage{subcaption}		
\usepackage{amssymb}
\usepackage{algorithm}		
\usepackage{algpseudocode}  
\usepackage{xcolor}			

\algrenewcommand\algorithmicindent{1.0em} 

\title{Multi-Stage Reinforcement Learning for Object Detection}

\author{
	Jonas König\\
	Paderborn University\\
	Paderborn, Germany\\
	\texttt{jonaskoe@mail.upb.de}\\
	\And
	Simon Malberg\\
	Paderborn University\\
	Paderborn, Germany\\
	\texttt{malberg@mail.upb.de}\\
	\And
	Martin Martens\\
	Paderborn University\\
	Paderborn, Germany\\
	\texttt{martensm@mail.upb.de}\\
	\And
	Sebastian Niehaus\\
	AICURA medical GmbH\\
	Berlin, Germany\\
	\texttt{sebastian.niehaus@aicura-medical.com}\\
	\And
	Artus Krohn-Grimberghe\\
	Lytiq GmbH\\
	Paderborn, Germany\\
	\texttt{artus@lytiq.de}\\
	\And
	Arunselvan Ramaswamy\\
	Paderborn University\\
	Paderborn, Germany\\
	\texttt{arunselvan.ramaswamy@uni-paderborn.de}\\
}

\begin{document}
\maketitle

\begin{abstract}
We present a reinforcement learning approach for detecting objects within an image.
Our approach performs a step-wise deformation of a bounding box with the goal of tightly framing the object.
It uses a hierarchical tree-like representation of predefined region candidates, which the agent can zoom in on.
This reduces the number of region candidates that must be evaluated so that the agent can afford to compute new feature maps before each step to enhance detection quality.
We compare an approach that is based purely on zoom actions with one that is extended by a second refinement stage to fine-tune the bounding box after each zoom step.
We also improve the fitting ability by allowing for different aspect ratios of the bounding box.
Finally, we propose different reward functions to lead to a better guidance of the agent while following its search trajectories.
Experiments indicate that each of these extensions leads to more correct detections.
The best performing approach comprises a zoom stage and a refinement stage, uses aspect-ratio modifying actions and is trained using a combination of three different reward metrics.
\end{abstract}


\section{Introduction}
\label{sec:introduction}

For humans and many other biological systems, it is natural to extract visual information sequentially \cite{Itti2005}. 
Even though non-biological systems are very different, approaches that are inspired by biological systems often achieve good results \cite{mathe2016, bellver2016}. 
In traditional computer vision, brute force approaches like sliding windows and region proposal methods are used to detect objects by evaluating all the proposed regions and choose the one or the ones where it is likely to fit an object \cite{uijlings2013}, \cite{zitnick2014}, \cite{girshick2015}. 
We propose a sequential and hierarchical object detection approach similar to the ones proposed by \cite{bellver2016} and \cite{caicedo2015}, which is refined after every hierarchical step. Like biological visual systems, the next looked at region depends on the previous decisions and states.\\
Our approach uses deep reinforcement learning to train an agent that sequentially decides which region to look at next. 
In our most sophisticated approach, the agent repetitively acts in two alternating action stages, zooming and moving, until it assumes to adequately frame an object with a tightly fitting bounding box. 
The actions allow to deform the bounding box enough to reach many bounding boxes of different sizes, positions and shapes. 
Because of the hierarchical zooming process, only a small number of region candidates must be evaluated, so that it is affordable to compute new feature maps after bounding box transformations, like in \cite{bellver2016}, instead of cropping from a single feature map.
This results in higher spatial resolutions of the considered regions, which are therefore more informative and allow for better decision making of the agent. 
We consider multiple reward functions using different metrics to better guide the training process of the agent compared to using the Intersection-over-Union. 
With these metrics the quality of a state is evaluated and rewarded not only with respect to the fit but also to the future potential of reaching a valid detection.
This leads to a speed-up in the training process and results in better detection quality.
\section{Related Work}
\label{sec:related_work}

Many works have already shown that reinforcement learning can be a powerful tool for object detection, without the need of object proposal methods like Selective Search \cite{uijlings2013} and Edge Boxes \cite{zitnick2014}. 
\cite{caicedo2015} casts the problem of object localization as a Markov decision process in which an agent makes a sequence of decisions, just as our approach does. 
They use deep Q-learning for class-specific object detection which allows an agent to stepwise deform a bounding box in its size, position and aspect ratio to fit an object. 
A similar algorithm was proposed by \cite{maicas2017}, which allowed for three-dimensional bounding box transformations in size and position to detect breast lesions. 
This approach reduced the runtime of breast lesion detection without reducing the detection accuracy compared to other approaches in the field.

Another approach is to perform a hierarchical object detection with actions that only allow to iteratively zoom in on an image region as shown by \cite{bellver2016}.
This does not allow for changes to the aspect ratio or movements of the bounding box without zooming. 
The main difference to the approaches before is that it imposes a fixed hierarchical representation that forces a top-down search which can lead to a smaller number of actions needed, but also leads to a smaller number of possible bounding box positions that can be reached. 
Because of the smaller number of actions, it becomes more affordable to extract high-quality descriptors for each region, instead of sharing convolutional features. 

\cite{mathe2016} proposed a method using a sequential approach related to ‘saccade and fixate’, which is the pattern of search in biological systems. 
Instead of deforming bounding boxes, the agent fixates interesting locations of an image and considers regions at that location from a region proposal method. 
The agent may terminate and choose one of the already seen bounding boxes for which it has the highest confidence of fitting the object or fixate on a new location based on already considered regions.
Our approach combines some of these approaches to create a 2-stage bounding box transformation process which uses zooming and allows for different aspect ratios and bounding box movements.
\section{Model}
\label{sec:model}

In this section, we define the components of our model and how it is trained.
We compare two general models.
The \emph{1-stage} model is based on the \emph{Image-Zooms} technique for hierarchical object detection as proposed by \cite{bellver2016}.
We introduce an extension to that model called \emph{2-stage}, which adds a way to perform refining movements of the bounding box.
\emph{2-stage} compensates for \emph{1-stage}'s drawback that only a very limited number of bounding boxes can be reached.
Additionally, we extend each model by actions to change the aspect ratio of the bounding box to improve the agent's ability to fit narrow objects.

\subsection{Markov Decision Process}
We use Q-learning to train the agent \cite{watkins1989}.
This reinforcement learning technique estimates the values of state-action pairs.
The agent chooses the action with the highest estimated future value regarding the current state.
We formulate the problem as a Markov decision process by defining states, actions and rewards, on which Q-learning can be applied.

\subsubsection{States}
The state contains the image features encoded by a VGG convolutional neural network \cite{simonyan2015} and a history vector that stores the last four actions taken by the agent.
The history vector is a one-hot vector as used by \cite{caicedo2015} and \cite{bellver2016}.
For the \emph{2-stage} model, we use two separate history vectors for the two stages.
The zoom state contains only the last four zoom actions and the refinement state only the last four refinement actions.
The refinement history vector is cleared after every zoom action.

\subsubsection{Actions}
\paragraph{1-stage}
Using the \emph{1-stage} model, the agent can choose between a terminal action and five zoom actions. The terminal action ends the search for an object and returns the current bounding box.
Zooming actions shrink the bounding box and thereby zoom in on one of five predefined subregions of the image:
Top-left, top-right, bottom-left, bottom-right and center.
Each zoom action shrinks the bounding box to 75\% of its width and height so that possible subregions overlap.
Such an overlapping is beneficial in situations where an object would otherwise cross the border between two possible subregions and could therefore not be framed without cutting off parts of the object.
By iteratively performing these zoom actions, the agent can further zoom into the image until it finally decides that the bounding box fits the object well.

\paragraph{1-stage-ar}
The \emph{1-stage-ar} model extends the \emph{1-stage} model by adding two zoom actions that allow to change the aspect ratio of the bounding box.
The bounding box can be compressed either in width or in height while holding the position of its center fixed.
For both actions, the bounding box is shrinked to 56.25\% of its width or height, respectively, to achieve an equal reduction of the bounding box area as compared to the other zoom actions.

\paragraph{2-stage}
The \emph{2-stage} model adds a second stage to the \emph{1-stage} model that performs five consecutive refinement steps after each zoom action.
In each refinement step, the agent can move the bounding box left, right, up or down by 10\% of its width or height, respectively, or decide to do nothing at all.

\paragraph{2-stage-ar}
The \emph{2-stage-ar} model analogously extends the \emph{1-stage-ar} model with a refinement stage.

\begin{figure}
	\includegraphics[width=\linewidth]{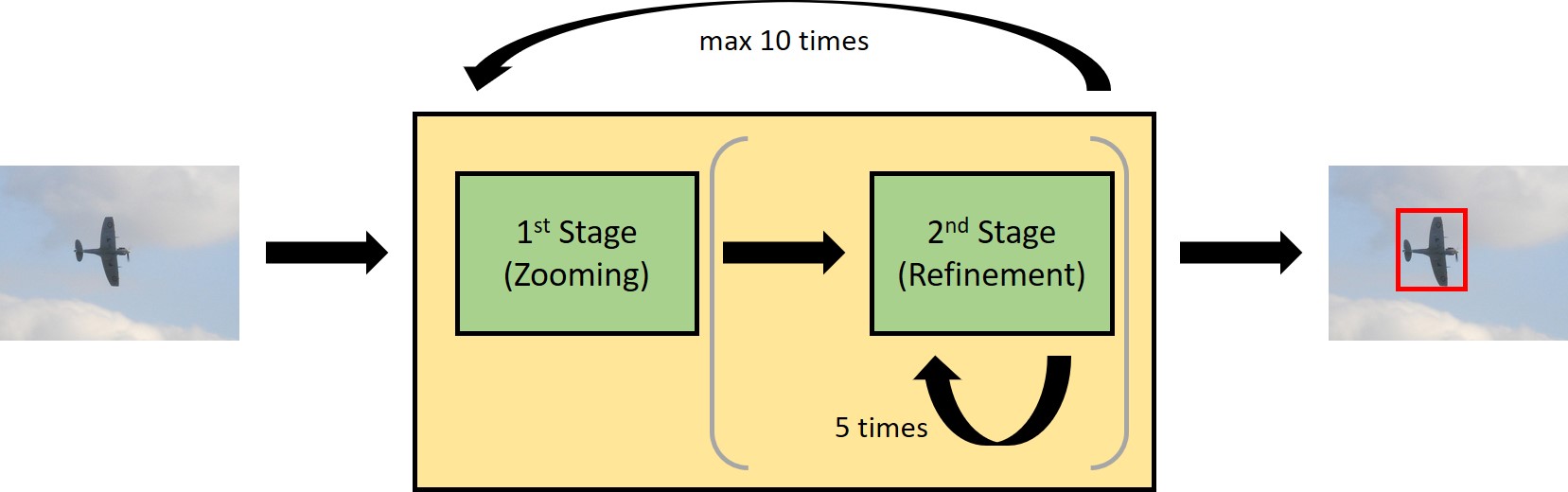}
	\caption{The agent can take up to ten zoom actions, which are determined by the first stage. For the \emph{2-stage} model, after each zoom step five refinement steps are taken, which are determined by the second stage.}
	\label{fig:procedure}
\end{figure}

\subsubsection{Reward}
\begin{figure}
\begin{center}
  \includegraphics[width=100mm]{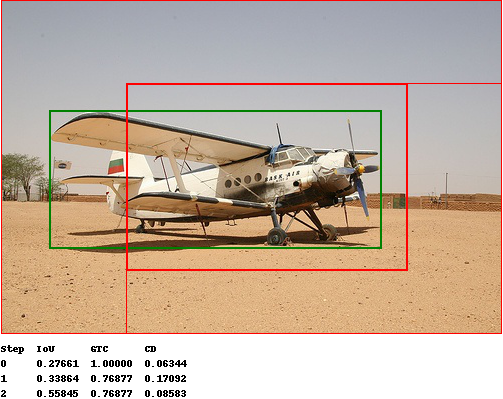}
\end{center}
  \caption{The first zoom increases the IoU by a small amount and therefore results in a positive reward, but also cuts off a large portion of the ground truth, which cannot be covered anymore in the future using the \emph{1-stage} approach.}
  \label{fig:fig1}
\end{figure}

The reward function scheme used is similar to the one proposed by \cite{caicedo2015, bellver2016}. 
They only used the Intersection-over-Union (IoU) as a metric to calculate the reward. 
Even though the correctness of a detection is evaluated using the IoU between the bounding box and the ground truth, using a zoom reward which is positive if a zoom increases the IoU and negative otherwise, does often not result in a favorable zoom action (see figure \ref{fig:fig1}). 
Therefore, more sophisticated rewards can be useful to guide the training process \cite{mataric1994}.
Multiple metrics were considered to be integrated in the reward function to better estimate the quality of a zoom step.
Two of them proved to be useful to increase the detection accuracy and lead to a faster convergence.

\paragraph{GTC}
The Ground Truth Coverage (GTC) is the percentage of the ground truth that is covered by the current bounding box. 
The main difference between IoU and GTC is, that the size of the bounding box has no direct impact on the GTC. 
If the whole ground truth is covered by the bounding box, the GTC is 1. 
If the bounding box covers nothing from the ground truth, the GTC is 0. 
The GTC metric can be used to put a higher weight on still covering a large portion of the ground truth after a zoom step.

\paragraph{CD}
The Center Deviation (CD) describes the distance between the center of the bounding box and the center of the ground truth in relation to the length of the diagonal of the original image. 
Therefore, the highest possible center deviation is asymptotically 1. 
Unlike IoU and GTC, a small CD is preferred. 
The CD metric can be used to support zooms in the direction of the ground truth center, which reduces the probability of zooming away from the ground truth.

\paragraph{Reward function}
Instead of using the IoU alone to evaluate the quality of a zoom state, we propose a more sophisticated quality function which additionally uses the introduced metrics to better evaluate the quality of zooms in a single equation.

\begin{equation}
r(b,g,d)= \alpha_1 iou(b,g) + \alpha_2 gtc(b,g) + \alpha_3 (1-cd(b,g,d))
\end{equation}

$r(b,g,d)$ estimates the quality of the current state and is calculated using bounding box $b$, ground truth $g$ and image diagonal $d$ as input weighted with factors $\alpha$.
The reward for zoom actions is calculated as follows:

\begin{equation}
R_z (b,b',g,d)=sign(r(b',g,d)-r(b,g,d))
\end{equation}

If the quality function value for the bounding box $b'$ after a zoom step is larger than the value for the bounding box $b$ before, the zoom is rewarded, otherwise it is penalized.
The terminal action has a different reward scheme which assigns a reward if the IoU of the region $b$ with the ground truth is larger than a certain threshold $\tau$ and a penalty otherwise. 
The reward for the terminal action is as follows:

\begin{equation}
R_t (b,g)=
\begin{cases}
+\eta & \text{if iou(b,g)} \ge \tau \\
-\eta & \text{otherwise}
\end{cases}
\end{equation}

In all our experiments, we set $\eta=3$ and $\tau=0.5$, as in \cite{caicedo2015, bellver2016}. 

\paragraph{Sigmoid}
In our experiments we also used a sigmoid function on each metric in the quality function to see if this nonlinear attenuation helps to improve the learning process.
A Gaussian attenuation was also considered but did not achieve a better performance. 
This results in a new quality function using sigmoids.

\begin{equation}
r(b,g,d)=\frac{\alpha_1}{1+e^{(\beta_1 (\gamma_1-iou(b,g)))}} + \frac{\alpha_2}{1+e^{(\beta_2 (\gamma_2-gtc(b,g)))}} + \frac{\alpha_3}{1+e^{(\beta_3(\gamma_3-(1-cd(b,g,d))))}}
\end{equation}

$\beta_i$ influences the strength of attenuation through the Sigmoid and $\gamma_i$ influences its centering. 
Different parameters were tested, and the best results were achieved with the following parameters.

\begin{equation}
\alpha_1,\alpha_2,\alpha_3=1 \quad	\beta_1,\beta_2,\beta_3=8 \quad	\gamma_1,\gamma_2,\gamma_3=0.5
\end{equation}

\paragraph{Refinement reward}
The refinement steps of the second stage are learned with a simpler reward function. 
This is because the only goal of that stage is to center the object within the bounding box. 
Therefore, only the CD metric is used in the reward function.

\begin{equation}
R_m (b,b',g,d)=
\begin{cases}
sign(cd(b,g,d)-cd(b',g,d)) & \text{if cd(b,g,d) $\neq$ cd(b',g,d)} \\
-1 & \text{if cd(b,g,d) = cd(b',g,d) while CD-decreasing refinements possible} \\
+1 & \text{if cd(b,g,d) = cd(b',g,d) while no CD-decreasing refinement possible}
\end{cases}
\end{equation}

Decreasing the Center Deviation is rewarded, increasing it is punished. 
If the agent does not move, the reward depends on the possibility to get a smaller CD. 
If the CD cannot be decreased using one of the moving refinements, then not moving is rewarded, otherwise it is punished.

\paragraph{Reward assignment}
In \emph{1-stage}, the zoom reward is calculated and assigned immediately after each zoom action.
In \emph{2-stage}, the reward for zooming is calculated and assigned after all five refinement actions have been taken to facilitate the coordination between the two stages.
The refinement reward is calculated and assigned directly after the refinement.

\subsection{Architecture}
\label{sec:architecture}

\begin{figure}
	\includegraphics[width=\linewidth]{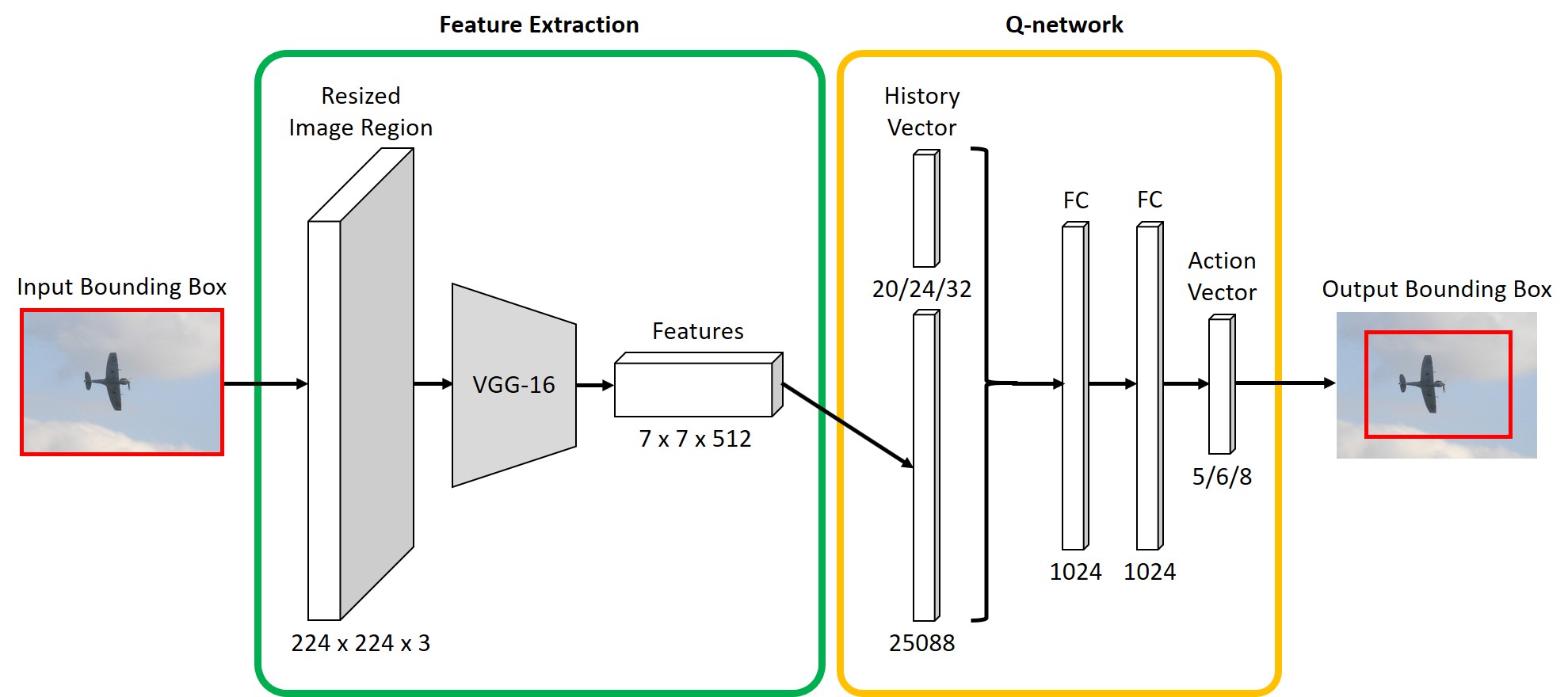}
	\caption{At each step, an image region as framed by the current bounding box is input to a VGG-16 convolutional neural network. The extracted image features concatenated with a history vector are then input to a Q-network. The Q-value outputs determine the next action to manipulate the bounding box.}
	\label{fig:stage_architecture}
\end{figure}

Both, the zoom stage and the refinement stage follow the layer architecture visualized in Figure \ref{fig:stage_architecture}.
At each step, they take the image and the current bounding box as input and output another bounding box, that should fit the object better.
The input image is cropped to the region determined by the bounding box and then resized to 224x224 pixels.
From this resized image region, a VGG-16 \cite{simonyan2015} extracts 512 7x7 sized maps of the region's image features.
These feature maps are flattened and concatenated with the agent's history vector containing the last four actions and passed through two fully connected layers with 1024 neurons each forming the Q-network.
Each fully connected layer is followed by a ReLU nonlinearity \cite{nair2010} and trained with 20\% dropout \cite{srivastava2014}.
The Q-network's output is a vector of Q-values containing one estimation per possible action.
The agent chooses the action with the highest Q-value and transforms the bounding box accordingly.
The image region inside the transformed bounding box serves as new input for the next step.

\subsection{Training}
In this section, we describe our approach for training the model and the hyperparameters used.

\paragraph{Optimization}
All models weights are initialized randomly from a normal distribution and optimized by Adam \cite{diederik2014} with a learning rate of $10^{-6}$.

\paragraph{Epsilon-greedy policy}
We balance exploration against exploitation using an epsilon-greedy policy.
Trainings start by taking random actions only ($\epsilon=1$) and linearly increase the likelihood of taking actions based on the learned policy until $\epsilon=0.1$, meaning 10\% of actions are random and 90\% are based on the learned policy.
For both models, $\epsilon$ decreases by 0.1 after each epoch.

\paragraph{Forced termination}
States in which the IoU is higher than 0.5 are relatively scarce when taking random actions.
Nevertheless, to learn the terminal action of the zoom model quickly, we enforce taking the terminal action whenever the IoU exceeds 0.5 as recommended by \cite{bellver2016}.

\paragraph{Experience replay}
In both the zoom stage and the refinement stage, consecutive states are very correlated, because a large portion of the considered image region is shared by both states.
To train the models with independent experiences, we apply an experience replay by storing experiences in a rolling buffer and training the model on a random subset of these stored experiences after each step.
For both stages, the size of the experience buffer is 1,000 experiences of which 100 are sampled randomly for the batch training.

\paragraph{Reward discounting}
Our Q-function considers future rewards with a discount factor of 0.9.
\section{Experiments}
\label{sec:experiments}

For a quantitative and qualitative comparison of different variations of our previously defined models, we conduct experiments in which we train our agent on one set of images and test it against another set of images.
To allow for a comparison with \cite{bellver2016}, we have used images of airplanes from the PASCAL VOC dataset \cite{pascal2010}.
Our agent has been trained on the 2012 trainval dataset and evaluated against the test dataset from 2007.

The performance of each model is measured by its true positive (TP) rate.
We consider a detection a true positive when the Intersection-over-Union of the predicted bounding box with the ground truth bounding box exceeds 0.5, as defined by the Pascal VOC challenge \cite{pascal2010}.
To measure the convergence over the course of multiple epochs of training, the models are evaluated against the test set after every tenth epoch of training.
For the \emph{1-stage} models, we conduct four runs for each model-reward combination and average the results.
For the \emph{2-stage} models, we take the average of three runs due to increased runtime requirements.

\subsection{Experimental Procedure}
\paragraph{1-stage}
Our first series of experiments serves the purpose of establishing a well-performing zooming methodology.
For this, we examine the effect of different reward functions and the effect of adding aspect ratio zoom actions.
Each model has been trained for 200 epochs on a set of 253 images and tested on a disjunctive set of 159 images.
Convergence curves indicate that results do not improve significantly by training longer than 200 epochs.

\paragraph{2-stage}
In a second series of experiments, the influence of a second-stage refinement is evaluated.
Training builds on top of a \emph{1-stage-ar} model that is pre-trained with an IoU, GTC and CD reward, namely the best-performing first-stage weights that reached a result of 51 true positives after 200 epochs.
For a duration of 100 epochs, only the refinement model is trained.
Results indicate that during training the detection quality of the \emph{2-stage} model starts off worse than \emph{1-stage}'s final results but surpasses them at around 100 epochs of refinement training.
Finally, both networks, zoom and refinement, are trained simultaneously for 50 epochs so that an appropriate coordination can be learned.

\subsection{Results}
As previously hypothesized, a combination of Intersection-over-Union, Ground Truth Coverage and Center Deviation outperforms a reward that is purely based on the Intersection-over-Union.
Experimental results indicate that these additional reward metrics help the agent to decide on its next zoom action, especially in situations where multiple zooms promise similar improvements in IoU.
GTC helps to minimize cutting-off mistakes that cannot be reverted.
CD assists to maintain focus on the object and prevent missing it on the zoom trajectory.

No matter which reward, the \emph{1-stage-ar} approach achieves more correct detections than the simpler \emph{1-stage} approach.
The fitting potential for narrow objects can apparently be exploited well enough to justify the two additional actions that must be learned by the agent.
For the \emph{1-stage-ar} models, the positive effect of considering additional reward metrics is even stronger than for the \emph{1-stage} models.
We hypothesize that GTC is particularly effective when aspect ratio zooms are possible.
In situations where an aspect ratio zoom is the optimal action, all other zoom actions usually cut off a large portion of the object.
This might result in an IoU increase but a larger decrease in GTC, so that taking an action other than the necessary aspect ratio zoom results in punishment.
However, implementing these reward metrics using a sigmoid squashing function seems to have no significant impact on the performance, maybe even a small negative impact on the performance of \emph{1-stage}.

Extending the model with a second stage for refinement leads to a further improvement in true positive detections.
The refinement ability is especially helpful in two types of situations:
(1) when the first stage makes an unfavorable zoom decision that can be corrected by the second stage and (2) when no predefined zoom region fits the object adequately but the second stage can refine the bounding box position enough to correct this offset.

\begin{figure}
	\centering
	\begin{subfigure}{.33\textwidth}
		\centering
		\includegraphics[width=.9\linewidth]{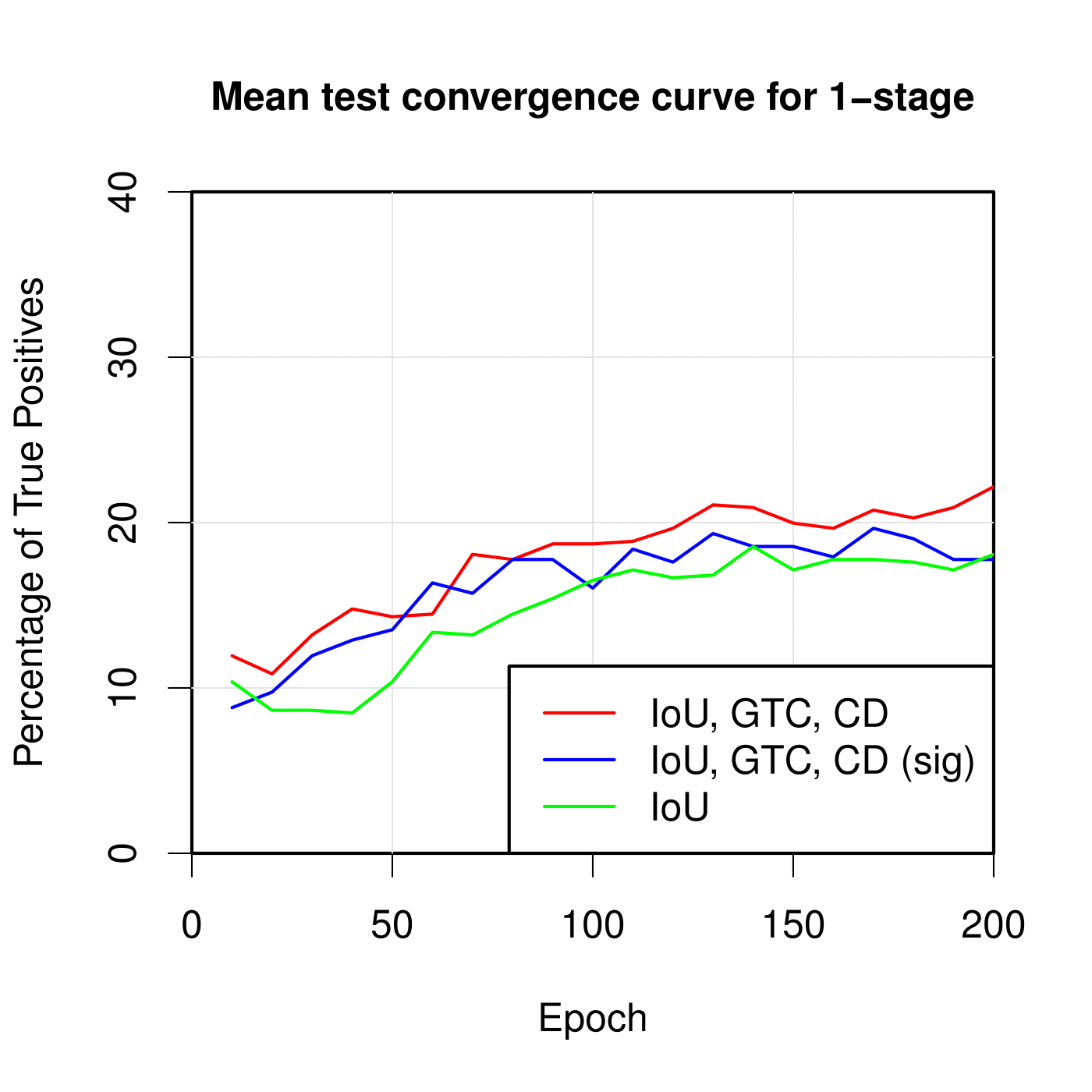}
		\caption{1-stage}
		\label{fig:results_1-stage}
	\end{subfigure}
	\begin{subfigure}{.33\textwidth}
		\centering
		\includegraphics[width=.9\linewidth]{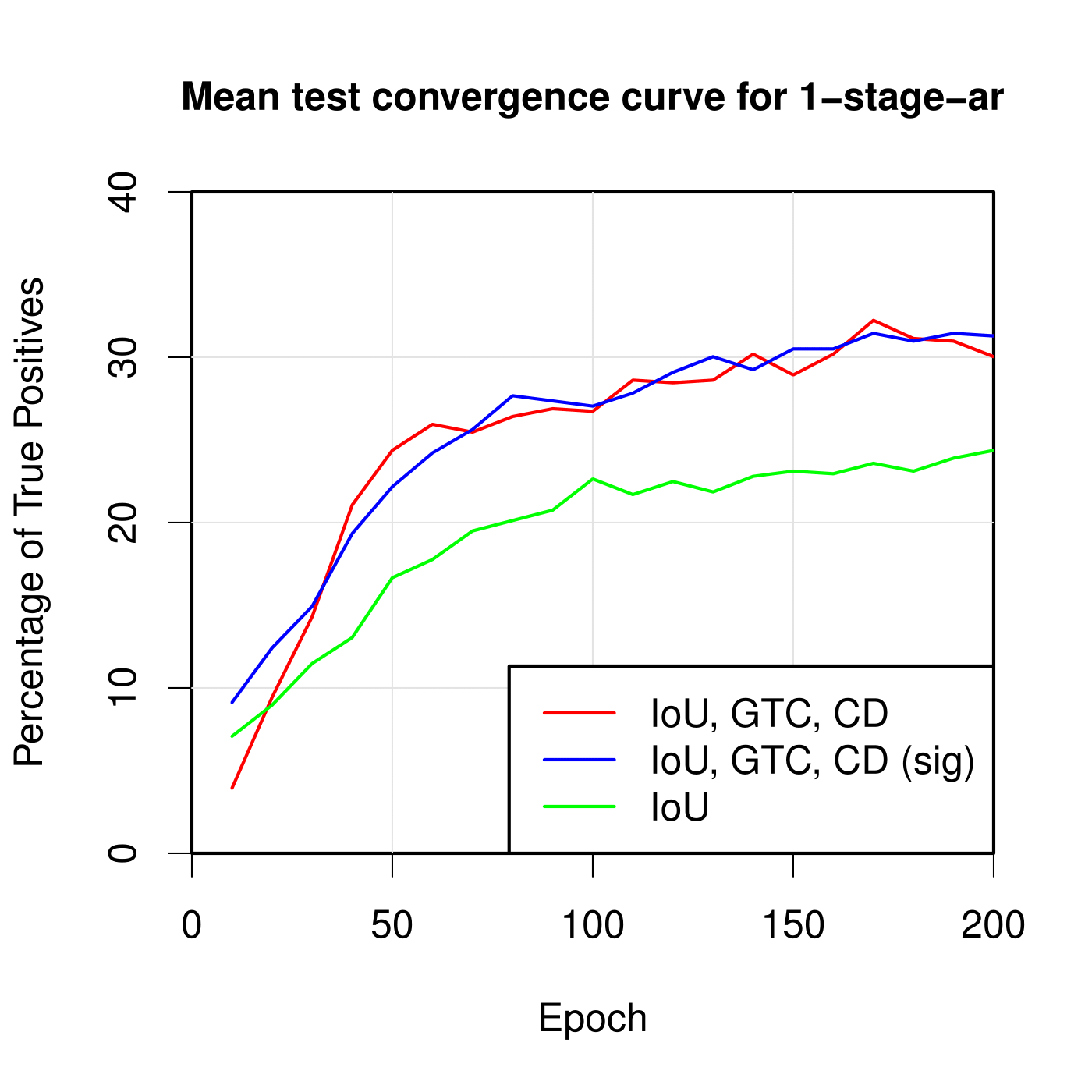}
		\caption{1-stage-ar}
		\label{fig:results_1-stage-ar}
	\end{subfigure}
	\begin{subfigure}{.33\textwidth}
		\centering
		\includegraphics[width=.9\linewidth]{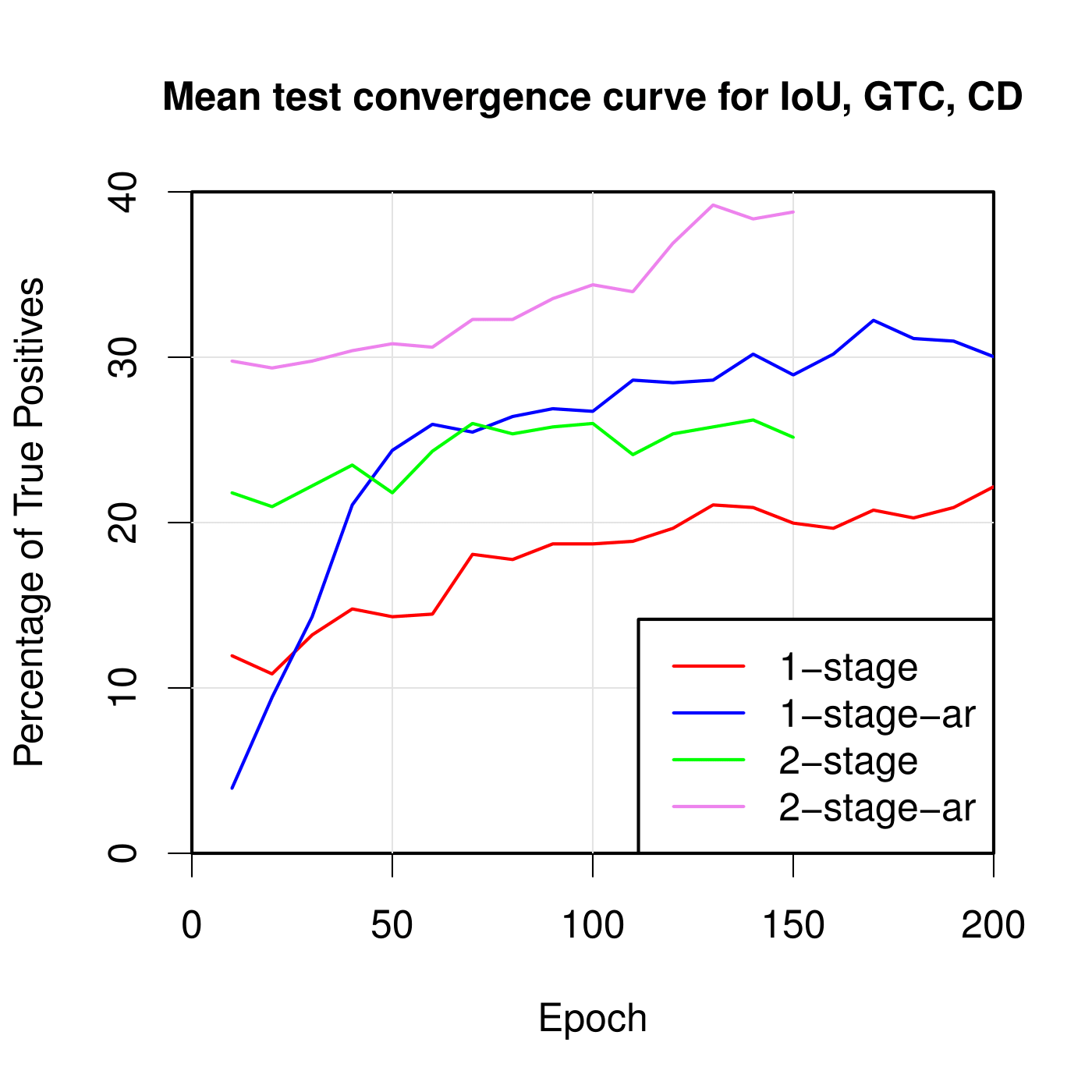}
		\caption{All models with IoU, GTC, CD}
		\label{fig:results_all}
	\end{subfigure}
	\caption{Convergence curves on the test set averaged over all runs}
	\label{fig:results}
\end{figure}

\begin{table}
	\label{tab:first_stage_results}
	\centering
	\begin{tabular}{|l|rrr|rrr|rrr|}
		\hline
		           & \multicolumn{3}{c|}{IoU} & \multicolumn{3}{c|}{IoU, GTC, CD} & \multicolumn{3}{c|}{IoU, GTC, CD (Sigmoid)}\\
		\hline
		Model      & TP  & FP  & FN  & TP  & FP  & FN  & TP  & FP  & FN  \\
		\hline
		\hline
		1-stage    &  28.75 & 102.50 &  27.75  & 35.25  & 90.75  & 33.00  & 28.25  & 99.00  & 31.75 \\
		\hline
		1-stage-ar &  38.75 &  92.00 &  28.25  & 47.75  & 89.50  & 21.75  & 49.75  & 86.00  & 23.25 \\
		\hline
		2-stage    &        &     &      & 40.00  & 95.00  &  24.00  & &     &    \\
		\hline
		2-stage-ar &        &     &      & 61.67  & 89.67  &  7.67  & &     &    \\
		\hline
	\end{tabular}
	\caption{Average results of all runs measured by their true positives (TP), false positives (FP) and false negatives (FP)}
\end{table}
\section{Conclusion}
\label{sec:conclusion}

This paper has presented a reinforcement learning approach that detects objects by acting in two alternating action stages.
The approach is based on a hierarchical object detection technique that makes it affordable to calculate feature maps of the image before each step.
This leads to a large variety of feature maps that will be extracted from each image during training, so that image data is already augmented by the model itself, which improves results especially on small data sets.
By adding actions that change the aspect ratio of the bounding box and by refining the bounding box after each zoom step, our model is capable of detecting 75\% more objects than without these supplementary actions.
Additionally, our experiments indicate that the number of correct detections can be increased by training the agent with reward metrics that are more informative about the quality and future potential of current states, namely Ground Truth Coverage and Center Deviation.
Combining aspect ratio zooms, refining movements and enhanced reward metrics, the detection rate could be nearly doubled in our experiments as compared to the pure zooming approach.
When taking into account that we did not train our \emph{2-stage-ar} model until convergence, one may expect even higher detection rates with more training.

Subsequent optimization of model parameters and fine-tuning of the reward weights should be considered to increase the number of correct detections even further.
Also, the detection accuracy might improve when the agent may choose different step sizes of zooming, aspect ratio changes and refining movements.

Admittedly, it is questionable if the presented approach may be labeled as traditional reinforcement learning.
Normally the ground truth is not known or not used to train the agent, which is not the case with our training method.
Instead, we use the ground truth annotation to shape our rewards and train the agent.
Therefore, our approach is supervised and uses reinforcement learning techniques at the same time.

\newpage
\bibliographystyle{unsrt}
\bibliography{references}

\newpage
\appendix

\section{Algorithm}
\begin{algorithm}[H]
	\caption{Training procedure (the blue lines are only necessary for the \emph{2-stage} models)}
	\label{alg:training}
	\begin{algorithmic}[1]
		\State $vggModel \gets loadVggModel()$
		\State $zoomModel \gets loadZoomModel()$
		\State $zoomExp \gets initializeZoomExpReplay()$
		\color{blue}
		\State $refModel \gets loadRefModel()$
		\State $refExp \gets initializeRefExpReplay()$
		\color{black}
		\For{$epoch \gets 1 $ \textbf{to} $maxEpochs$}
			\ForAll{$image$ \textbf{in} trainingset \textbf{with} one object of interest}
				\State $groundTruth \gets loadGroundTruth(image)$
				\State $boundingBox \gets initializeBoundingBox()$
				\State $historyVectorZoom \gets initializeHistoryVectorZoom()$
				\State $zoomState \gets getZoomState(vggModel,image, boundingBox, historyVectorZoom)$
				\State $zoomStep \gets 0$
				\State $terminated \gets False$
				\While{\textbf{not} $terminated$ \textbf{and} $zoomStep < 10$}
					\State $prevZoomBoundingBox \gets boundingBox$
					\State $prevZoomState \gets zoomState$
					\State $zoomAction \gets chooseZoomAction(zoomModel, zoomState)$
					\State $zoomAction \gets checkForcedTermination(boundingBox, groundTruth, zoomAction)$
					\If{$zoomAction \neq$ terminal action}
						\State $boundingBox \gets performZoomAction(boundingBox, zoomAction, image)$
						\color{blue}
						\State $historyVectorRef \gets initializeHistoryVectorRef()$
						\State $refState \gets getRefState(vggModel,image, boundingBox, historyVectorRef)$
						\For{$refStep \gets 1$ \textbf{to} $5$}
							\State $prevRefBoundingBox \gets boundingBox$
							\State $prevRefState \gets refState$
							\State $refAction \gets chooseRefAction(refModel, refState)$
							\State $boundingBox \gets performRefAction(boundingBox, refAction, image)$
							\State $historyVectorRef \gets updateHistoryVectorRef(historyVectorRef, refAction)$
							\State $refReward \gets calculateRefReward(boundingBox, prevRefBoundingBox, groundTruth, image)$
							\State $refState \gets getRefState(vggModel,image, boundingBox, historyVectorRef)$
							\State $refExp \gets memorizeRefExp(prevRefState, refState, refStep, refReward, refAction)$
						\EndFor
						\color{black}
					\EndIf
					\State $zoomStep \gets zoomStep +1$
					\State $terminated \gets zoomAction =$ terminal action
					\State $historyVectorZoom \gets updateHistoryVectorZoom(historyVectorZoom, zoomAction)$
					\State $zoomReward \gets calculateZoomReward(boundingBox, prevZoomBoundingBox, groundTruth, image)$
					\State $zoomState \gets getZoomState(vggModel,image, boundingBox, historyVectorZoom)$
					\State $zoomExp \gets memorizeZoomExp(prevZoomState, zoomState, zoomReward, zoomAction)$
					\State $fitZoomModel(zoomModel, zoomExp)$
					\color{blue}
					\State $fitRefModel(refModel, refExp)$
					\color{black}
				\EndWhile
			\EndFor
		\EndFor
	\end{algorithmic}
\end{algorithm}

\newpage
\section{Evaluation overview}
\begin{figure}[H]
	\begin{center}
		\includegraphics[width=.9\linewidth]{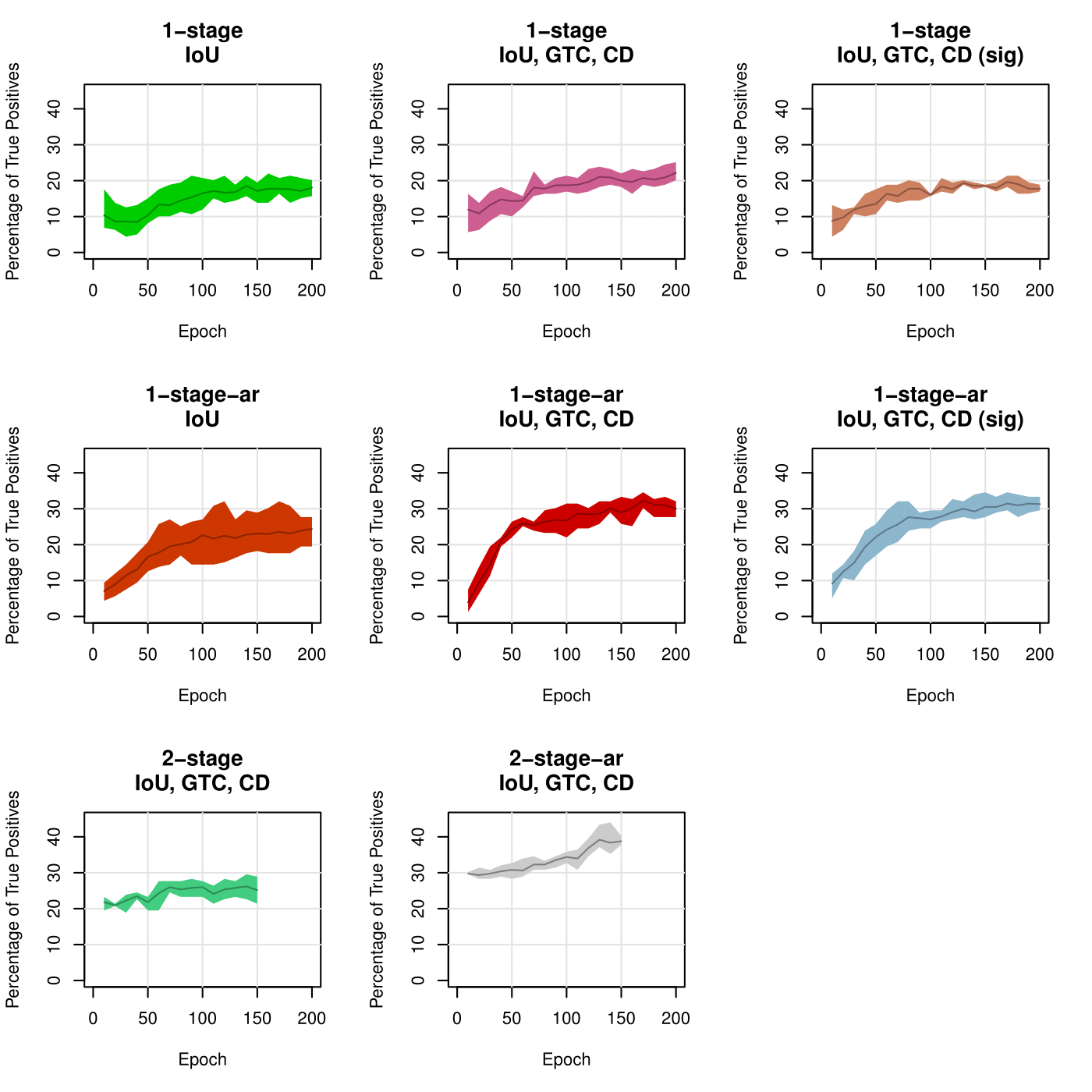}
	\end{center}
	\caption{Convergence curves of the evaluated approaches. The colored areas visualize the min-max value ranges, the lines represent the mean values.}
	\label{fig:eval_overview}
\end{figure}
\end{document}